\documentclass[letterpaper, 10 pt, conference]{ieeeconf}  

\IEEEoverridecommandlockouts                              

\usepackage{graphicx}

\usepackage{tabularray}
\usepackage{cite}
\usepackage{color}
\usepackage{tabularray}
\usepackage{amsmath} 
\usepackage{amssymb}  

\title{\LARGE \bf
Cross-Domain Transfer Learning using Attention Latent Features for Multi-Agent Trajectory Prediction
}

\author{Jia Quan Loh$^{1,*}$, Xuewen Luo$^{2,*}$, Fan Ding$^{2,*}$, Hwa Hui Tew$^{2,*}$, Junn Yong Loo$^{2,\dagger}$, \\Ze Yang Ding$^{1}$, Susilawati Susilawati$^{1}$, Chee Pin Tan$^{1}$ 
\thanks{The work of Junn Yong Loo is supported by the Ministry of Higher Education Malaysia under the Fundamental Grant Scheme (FRGS) G-M010-MOH-000206, and Monash University under the SIT Collaborative Research Seed Grants I-M010-SED-000242.}
\thanks{$^{1}$The authors are with the School of Engineering, Monash University Malaysia.
$^{2}$The author is with the School of Information Technology, Monash University Malaysia.}
\thanks{$^{*}$The authors contributed equally to this work.
$^{\dagger}$Corresponding author.}
}

\begin{document}

\maketitle
\thispagestyle{empty}
\pagestyle{empty}

\begin{abstract}
With the advancements of sensor hardware, traffic infrastructure and deep learning architectures, trajectory prediction of vehicles has established a solid foundation in intelligent transportation systems. However, existing solutions are often tailored to specific traffic networks at particular time periods. Consequently, deep learning models trained on one network may struggle to generalize effectively to unseen networks. To address this, we proposed a novel spatial-temporal trajectory prediction framework that performs cross-domain adaption on the attention representation of a Transformer-based model. A graph convolutional network is also integrated to construct dynamic graph feature embeddings that accurately model the complex spatial-temporal interactions between the multi-agent vehicles across multiple traffic domains. The proposed framework is validated on two case studies involving the cross-city and cross-period settings. Experimental results show that our proposed framework achieves superior trajectory prediction and domain adaptation performances over the state-of-the-art models.
\end{abstract}

\section{INTRODUCTION}
In the realm of intelligent transportation, the landscape of vehicle trajectory prediction has evolved significantly and gained immense traction in intelligent transportation systems, spanning applications from vehicle design to traffic forecasting and traffic control \cite{1}. This transformation is driven by advancements in vehicle sensor hardware and traffic infrastructure. Consequently, this allows the acquisition of high fidelity sensory and positional data of multiple vehicles, which are crucial for data-driven modelling of the complex spatial-temporal interactions between vehicles in a multi-agent traffic network \cite{2}. In particular, an accurate forecast of the future vehicular trajectories allows a ego vehicle to plan its optimal navigational route within the network and alleviate traffic issues such as congestion and accidents.

Recent advances in the deep learning paradigm have tremendously enhanced vehicular trajectory prediction in many traffic networks such as arterial roads, boulevards, and interstate highways. A noteworthy example is the graph-based interaction-aware trajectory prediction (GRIP) model \cite{3} which incorporates graph convolutional neural network (GCN) and a recurrent encoder-decoder architecture. The GRIP model exploited graph representation to model complex spatial inter-agent interactions and the sequential encoding modules in recurrent neural network (RNN) to model temporal correlation across the vehicle trajectories. Apart from that, attention mechanism of the Transformer networks \cite{4,5} has been harnessed to effectively model time-series data. The attention mechanism involves a global treatment of the time-evolving trajectory as a unified sequence, thus mitigating the deficiency of RNNs in retaining long-term temporal dependencies in long vehicle trajectory \cite{6,7}.

Nevertheless, deep trajectory prediction models are often tailored to the available training data, which could be collected on a particular traffic configuration such as time period or fixed location. This adherence to a certain traffic domain inhibits the model from effectively generalizing its prediction to unseen traffic networks \cite{8}. Current domain adaptation approaches for trajectory prediction predominantly rely on semi-supervised techniques 
\cite{10}. These approaches serve as effective means to transfer existing knowledge within vision models pertaining to road geometry and topology, and in kinematic models for understanding driver maneuvering behaviors. For example, Xu et al. \cite{11} have tackled the domain shift challenge by employing domain adaptation techniques such as similarity losses between source and target domains for distribution alignment in the context of pedestrian trajectory prediction. Nevertheless, a shift in geographical location and distinct traffic conditions could make these models ineffective when applied to comprehensive system-wide network. Moreover, traffic dynamics evolve over time and thus the validity of a trained model would be limited to the given temporal window. 

Motivated by these difficulties, in this paper, we propose a novel sequence-to-sequence graph Transformer-based model (namely Graph Embedded Transformer) to learn the spatial-temporal features of multi-agent trajectory data. Our proposed framework utilizes the embedding capabilities of GCN to help model the spatial features of multi-agent trajectories, while the Transformer performs temporal modelling of the trajectory sequence. To address the issue of co-variate domain shifting due to the differences in traffic distributions in different locations or periods, we introduce a domain adaptation training strategy on top of the 
spatial-infused attention embedding of the Transformer encoder to 
adapt the model's attention across multiple traffic domains.

To the best of our knowledge, 
this is the first work that investigates the feasibility of a domain-adaptable graph Transformer-based framework 
on generalizing trajectory prediction from source domain to target domain with limited training data.
Our contributions are highlighted as follows:
\begin{itemize}
        \item We introduce 
        the Graph Embedded Transformer, which first integrates
        a Graph Convolutional Network to construct spatial-aware non-Euclidean input embeddings for the Transformer \cite{13}. The Transformer's encoder module then perform temporal encoding of the spatial input embeddings to encode historical trajectory. 
        The Transformer's decoder module subsequently generates future trajectory based on the encoder's latent features.
        \item A domain adversarial training strategy is incorporated on top of the Transformer's encoder module is to achieve cross-domain transfer learning across traffic domains. In particular, the latent representation of the encoders composed of its spatially-infused attention embeddings are input into a discriminatory layer to minimize the statistical discrepancy between the latent space in different domains.
        \item The efficacy of the proposed Graph Embedded Transformer is validated against the state-of-the-art vehicle trajectory prediction models.
        Comparative results on the NGSIM-I80 and NGSIM-US101 datasets 
        show that our proposed model 
        achieves superior accuracy 
        across the source and target domains, thus indicates effective domain generalization.
\end{itemize}  

The rest of this paper is organized as follows. In Section II, we briefly discuss related work followed by the problem formulation in Section III. In Section IV, we describe our base architectures used and implementation details for domain adaptation. We report our experimental results in Section V. Finally, we conclude this paper in Section VI.

\begin{figure*}[t]
    \centering
    \includegraphics[width=0.95\textwidth]{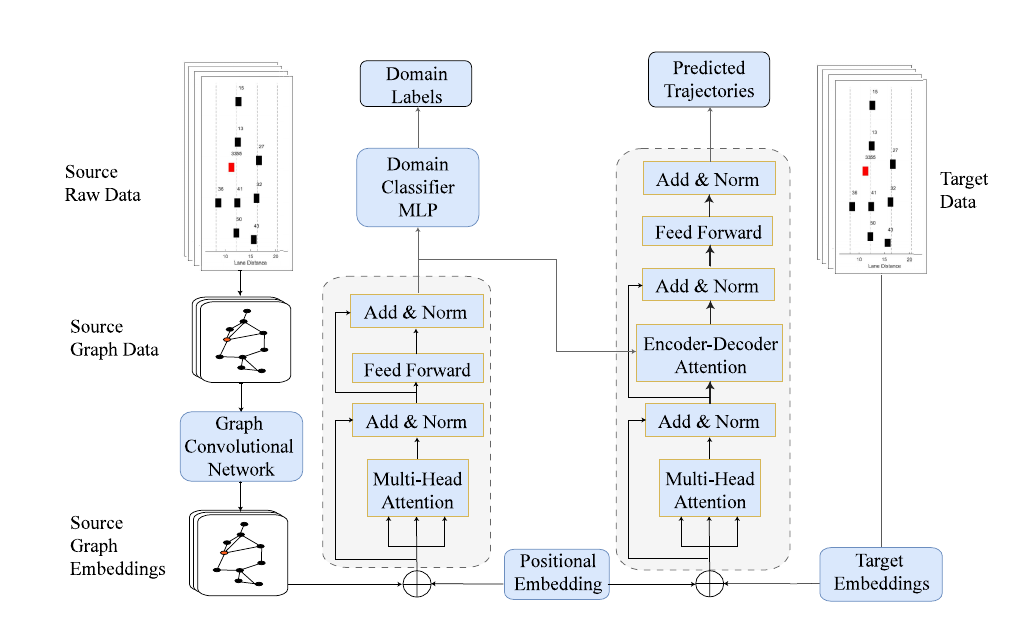}
    \caption{Graph Embedded Sequence-to-Sequence Transformer integrated with Domain Adaptation. The target is separated into (a) Training Mode: Last segment from Source + Output shifted right and (b) Inference Mode: Last Segment from Source + Iterative Predictions shifted right.}
    \label{trans}
\end{figure*}

\section{RELATED WORK}

\subsection{Trajectory Prediction Architecture}
In earlier studies, the utilization of Recurrent Neural Networks (RNNs) in trajectory prediction architecture was prominent, primarily owing to their adeptness in capturing temporal features and modeling sequential data \cite{16,17}. Subsequently, Long Short-Term Memory (LSTM) \cite{12} was introduced, and it is proved effective in learning long-term temporal dependencies in traditional RNNs \cite{16}. The LSTM proved particularly suitable for capturing driving experiences characterized by extended temporal delays within the sequence, as showcased by the works \cite{18,19} which incorporated LSTM or LSTM encoder-decoder models into their vehicle intention prediction frameworks. Furthermore, the Environment-Attention architecture \cite{20} adopted an LSTM encoder-decoder approach to directly forecast the future positions of vehicle trajectories. It is worth noting, however, that LSTM's prediction speed is hindered by the need for individual analysis of each time frame in the trajectory sequences of surrounding vehicles \cite{21} , which lengthens computation speed and increases network size as the sequence length increases.
To mitigate computational demands, an alternative to LSTM, namely the Gated Recurrent Unit (GRU), was introduced by Li et al. \cite{3} into their existing Graph-based Interaction-aware Trajectory Prediction (GRIP) model. GRU offers the advantage of having lesser model parameters while retaining/without compromising model performance.

In recent works, there has been a notable shift towards the adoption of Transformer-based or attention-based networks, which have demonstrated their proficiency in temporal analysis of extended input sequences through their capacity to mimic cognitive attention processes \cite{4,5,7,tf1}. These networks have emerged as a deep learning solution for effective modeling and extraction of temporal features from sequential data. To achieve this, attention mechanisms have been incorporated into the feature space encoding to prioritize essential node features within specific time frames.
For example, \cite{4,7} have also harnessed the multi-headed attention mechanism during the encoding of trajectory histories in their Transformer networks. This approach enhances the network's stability during training while enabling accurate long-term predictions with reduced inference time. This makes Transformer networks a compelling choice for real-time applications, including autonomous driving.
In conjunction with temporal sequence processing, the incorporation of spatial interaction modeling stands as a pivotal element within trajectory prediction models. 
In the study conducted by Mo et al. \cite{cnnlstm}, Convolutional Neural Networks (CNNs) were incorporated into the framework for single-agent vehicular trajectory prediction. This integration aimed to imbue interaction awareness within the LSTM network, yielding favorable outcomes in trajectory prediction. Notably, the proposed approach demonstrated efficacy, particularly in scenarios characterized by prominent lane-changing dynamics within the traffic environment.

Recently, Graph Neural Networks (GNNs) have gained popularity as a compelling alternative for modeling spatial interactions among vehicles. Notably, two prevalent models, namely Graph Convolutional Networks (GCNs) \cite{gcn1} and Graph Attention Networks (GANs) \cite{gcn2}, have been adopted for deep learning on graph structures.
In particular, Li et al. \cite{3} and Sheng \cite{gcn3} have incorporated GCNs into their architectures to extract spatial features in conjunction with various forms of Recurrent Neural Networks (RNNs) to capture temporal features. Additionally, Zhang et al. \cite{gcn4} introduced Graph Attention Transformer (GAT) to incorporate importance and attention into graph edges to effectively model spatial features. Notably, Zhang et al. \cite{gcn4} extended their graph-based approach by passing satellite maps relative to the ego vehicle's location through  CNNs into their architecture. This augmentation facilitates obstacle extraction, with the extracted features from the CNN being subsequently encoded into the graph structures of the GAT block. Consequently, their architecture proficiently captures inter-vehicle spatial and temporal interactions, as well as interactions between vehicles and road obstacles during training.

\subsection{Cross-Domain Adaptation Training}

Trajectory data originating from diverse geographic locations belongs to distinct domains \cite{11}. A model trained on a source domain, when applied to a different target domain, introduces domain shift which compromises its ability to generalize effectively to the new domain \cite{22}. Domain adaptation methods hold significant promise for mitigating this by minimizing the disparity in the structural relationships among time series variables across multiple domains \cite{23}.

A prevalent strategy in domain adaptation involves constructing latent representations within a feature space, where feature vectors with similar distributions are positioned closer to each other\cite{24}. Notably, \cite{23} expanded domain adaptation techniques for time-series forecasting within Transformer networks by introducing the concept of shared attention among a model that is trained on a multi-domain data set.

\section{PROBLEM FORMULATION}
We define the trajectory prediction problem as a forecasting task of the future positions of all vehicles within a given traffic scene. Our prediction horizon spans a 5-second interval, drawing upon trajectory histories encompassing up to 3 seconds. The models are fed with input data denoted as $X$, which represents the trajectory histories of all vehicles within the scene over a time span of $t_{h}$ time steps, as follows:
\begin{equation}
X =\big[ p^{1}, p^{2},..., p^{t_h} \big],
\end{equation}
where
\begin{equation}
p^t=\big[ \, (x_0^t,y_0^t ),(x_1^t,y_1^t ),\dots,(x_n^t,y_n^t) \, \big]
\end{equation}
are the coordinates of all observed vehicles for a traffic scene at time $t$, and $n$ is the number of observed vehicles. 
The output $Y$ of the models are the predicted positions of all the observed objects from time frames $t_h+1$ to $t_h+t_f$ in the future, as follows:
\begin{equation}
Y=\big[ p^{t_h+1}, p^{t_h+2},\dots, p^{t_h+t_f} \big].
\end{equation}
The objective function for predicting trajectories is defined using the Mean Squared Error (MSE) loss, as depicted below
\begin{equation}
\mathrm{MSE}\,(Y_i, \hat{Y}_i) = \frac{1}{N} \sum_{i=1}^{N} \big( Y_i - \hat{Y}_i \big)^2,
\end{equation}
where $N$ denotes the batch size (number of trajectories in a batch) used for training.

\section{PROPOSED METHODS}
This section introduces a domain adaptation approach that leverages the latent space representations of the Graph Embedded Transformer. This proposed approach aims to facilitate the transfer of trajectory prediction knowledge from a source domain, characterized by abundant data, to a target domain with limited data availability. A visual representation of the proposed model is depicted in Fig \ref{trans}.

\subsection{Input Representation} 
Given the observation of $n$ vehicles over a time span of $t_h$ seconds, the information is structured into a 3D array with dimensions ($t_h$ × $n$ × $f$) with $f=2$ number of features representing the $x$ and $y$ coordinates of each vehicle. Subsequently, the data is subjected to a normalization process, constraining its range to $[-1, 1]$.

\subsection{Trajectory Prediction Model}
In this subsection, we provide a concise overview of the architectures which serves as the benchmark models as well as the baseline of our proposed framework.

\subsubsection{Recurrent Neural Network}
We employ an LSTM auto-encoder network wherein each time step inputs data into the encoder LSTM. The hidden features extracted by the encoder block, alongside the vehicle coordinates from the previous time step, are subsequently fed into the decoder LSTM to forecast the positional coordinates for the current time step. This decoding process iterates until the model provides predictions for a specified future time horizon denoted as $t_f$.
To introduce interaction awareness, we incorporate CNNs as utilized by \cite{cnnlstm} to form the CNN-LSTM network. We also implement the GRIP architecture introduced by \cite{3} by incorporating GCNs into the LSTM network.

\subsubsection{Attention Networks}
We employ a Sequence-to-Sequence Transformer model, as detailed in \cite{13}, which incorporates multi-headed self-attention within its encoder block and decoder block. 
The input trajectory, spanning $t_h$ time steps, is concatenated to serve as the source input. During the training phase, the target input consists of the right-shifted output trajectory, which is preceded by the final time frame from the input trajectory. In the inference phase, the target input is iteratively derived from the predicted trajectory at each output time frame until all the $t_f$ time frames have been processed.
In view of the promising performance of Graph Neural Networks in trajectory prediction \cite{3,4}, our proposed model introduces a Graph Convolution Block to capture spatial dependencies of the input data before feeding them into the Transformer. This architecture design yields the proposed model termed Graph Embedded Transformer, which serve as our baseline for incorporating domain adaptation.

\subsection{Domain Adversarial Training}
As highlighted in \cite{8,10,11}, the challenge of domain shift remains a persistent concern in cross-domain trajectory prediction. This issue can render trajectory prediction models ineffective when applied to different geographical locations or varying traffic scenarios. Iterative model training across diverse locations is impractical due to its substantial computational demands.
In light of these considerations, our approach seeks to adapt the models to both the source and target domains. We achieve this by extracting the latent space representation of each trajectory and directing it through an MLP discriminatory network, responsible for classifying the latent features into their respective domains. This transfer strategy facilitates the adjustment of the source domain's latent space to align with that of the target domain. Consequently, our primary objective is to minimize the dissimilarity between the latent features of the source and target domains, a goal achieved through the optimization of a binary cross-entropy (BCE) loss as follows:
\begin{equation}
L(a_i, \hat{a}_i) = - \frac{1}{N} \sum_{i=1}^{N} \left[ \, a_i \log(\hat{a}_i) + (1 - a_i) \log(1 - \hat{a}_i) \, \right]
\end{equation}
where $a_i \in \{0, 1\}$ is the corresponding domain label predicted by the domain classifier, which indicates if the latent attention of the Transformer's encoder belongs to the source or target domain. 

Motivated by the domain adversarial training (DAT) strategy \cite{Ganin}, the MSE and BCE losses are backpropagated to optimize the model parameters of the Transformer encoder module and the domain classifier, respectively.
Such an adversarial competition between the Transformer's encoder and the MLP classifier is expected to achieve a Nash equilibrium that minimizes the statistical discrepancy between the latent distributions and thus generalizes the learned latent space to both the source and target domains \cite{Chai}.
\definecolor{Gray}{rgb}{0.498,0.498,0.498}
\definecolor{Black}{rgb}{0,0,0}
\begin{table*}[t]
\caption{CROSS-CITY STUDY: COMPARISON BETWEEN RMSE OF THE PROPOSED DOMAIN ADAPTED FRAMEWORK AND BENCHMARK MODELS ON SOURCE DOMAIN (NGSIM-I80). RMSE IS RECORDED IN METERS.}
\label{t1}
\centering
\begin{tblr}{
  cells = {c},
  cell{1}{1} = {c=7}{},
  hline{1} = {-}{},
  hline{2-3,9-10} = {-}{Black},
  vline{7} = {2-9}{},
  vline{6} = {2-9}{}
}
\textbf{Source Domain (NGSIM-I80)}    &         &                 &          &             &                              &                                 \\
\textbf{Prediction Horizon (s)} & LSTM   & GRIP            & CNN-LSTM & Transformer & {
  GCN-
  \\Transformer
  } & {
  GCN-
  \\Transformer (DA)
  } \\
\textbf{0}                      & 6.1972  & \textbf{2.0942} & 3.8789   & 2.7825      & 2.7606                       & 2.5295                          \\
\textbf{1}                      & 7.3355  & \textbf{2.4305} & 4.1643   & 3.3973      & 3.3714                       & 2.7349                          \\
\textbf{2}                      & 8.6745  & 3.2804          & 4.5903   & 3.6273      & 4.0040                       & \textbf{2.9154}                 \\
\textbf{3}                      & 9.1429  & 3.7860          & 4.4240   & 4.2293      & 4.3508                       & \textbf{3.2654}                 \\
\textbf{4}                      & 9.5990  & 4.3883          & 4.4651   & 4.7011      & 4.7395                       & \textbf{3.7176}                 \\
\textbf{5}                      & 10.7344 & 5.3638          & 4.9250   & 4.7378      & 5.1302                       & \textbf{4.1254}                 \\
\textbf{Average}                & 8.6139  & 3.5572          & 4.4079   & 3.9126      & 4.0594                       & \textbf{3.2147}                 
\end{tblr}
\end{table*}

\begin{table*}[t]
\caption{CROSS-CITY STUDY: COMPARISON BETWEEN RMSE OF THE PROPOSED DOMAIN ADAPTED FRAMEWORK AND BENCHMARK MODELS ON TARGET DOMAIN (NGSIM-US101). RMSE IS RECORDED IN METERS.}
\label{t2}
\centering
\begin{tblr}{
  cells = {c},
  cell{1}{1} = {c=7}{},
  hline{1} = {-}{},
  hline{2-3,9-10} = {-}{Black},
  vline{7} = {2-9}{},
  vline{6} = {2-9}{}
}
\textbf{Target Domain (NGSIM-US101)}  &         &                 &          &             &                              &                                 \\
\textbf{Prediction Horizon (s)} & LSTM   & GRIP            & CNN-LSTM & Transformer & {
  GCN-
  \\Transformer
  } & {
  GCN-
  \\Transformer (DA)
  } \\
\textbf{0}                      & 6.7265  & \textbf{2.7626} & 4.2010   & 3.2595      & 3.2499                       & 2.9618                          \\
\textbf{1}                      & 7.5750  & \textbf{3.0435} & 4.3351   & 3.8049      & 3.9101                       & 3.1250                          \\
\textbf{2}                      & 8.4709  & 3.5861          & 4.4489   & 3.8604      & 4.1135                       & \textbf{3.1889}                 \\
\textbf{3}                      & 9.4912  & 4.1650          & 4.5522   & 4.5354      & 4.6013                       & \textbf{3.4771}                 \\
\textbf{4}                      & 10.5377 & 4.9373          & 4.8485   & 4.9593      & 5.1761                       & \textbf{3.9401}                 \\
\textbf{5}                      & 11.8979 & 5.7341          & 5.2815   & 5.1764      & 5.5377                       & \textbf{4.1877}                 \\
\textbf{Average}                & 9.1165  & 4.0381          & 4.6112   & 4.2658      & 4.4314                       & \textbf{3.4751}                 
\end{tblr}
\end{table*}

\section{EXPERIMENTS}

\subsection{Datasets}
The datasets are evaluated on two trajectory prediction datasets: NGSIM-I80 
and the NGSIM-US101 
dataset. Both datasets were captured at 10 Hz over 45 minutes and segmented into 15 minutes of mild, moderate and congested traffic conditions. These two datasets consist of trajectories of vehicles on real freeway traffic. Each trajectory is segmented into 8 second intervals, in which the first 3 seconds are used as the trajectory history and the remaining 5 seconds are the prediction ground truth. The data is down sampled for each segment by a factor of 2, i.e. 5 Hz as in \cite{3}. This provides 40 frames for each 8 second trajectory, such that $t_h$ spans 15 frames in 3 seconds and $t_f$ spans 25 frames in 5 seconds.

\subsection{Implementation Details}
The proposed methods are implemented using Python and the Pytorch deep learning library. The settings of model and training hyperparameters used are reported below.

\subsubsection{Input Representation}
In this paper, we process a traffic scene within 30 meters and located in the same lane or at the adjacent lanes of an ego vehicle. All vehicles within this region will be observed and predicted into the future.

\subsubsection{Model Hyperparameters}
The Graph Embedded Transformer uses four attention heads with a MLP of size of 2048. Two layers of Transformer encoders and decoders are used. The input embedding model is replaced with a Graph Convolution Network to aid in the spatial modelling of our proposed Graph Embedded Transformer.

\subsubsection{Domain Adaptation}
A simple MLP is used for the discriminatory network to classify the latent space of the Graph Embedded Transformer into their respective domains. 

\subsubsection{Optimization}
The model training comprises a regression task for the trajectory prediction and a classification task for the domain classification. The overall regression loss is the MSE and the domain classification loss is the BCE. The models are concurrently trained to minimize these losses.

\subsubsection{Training Hyperparameters}
The models are trained using the ADAM optimizer with a 0.0001 learning rate without annealing. Batch size is set to 128 for training.

\subsection{Domain Adaptation Case Studies}
Two cross-domain scenarios are considered in our case studies. In the source domain, a third of the trajectory data are used for testing. In the target domain, a half of the trajectory data are used for testing. A detailed breakdown is provided in the following subsections.

\subsubsection{Case Study 1 - Cross-city Transfer Learning}
We analyze the effects of latent space domain adaptation on the cross-city scenario by using trajectory data from both the NGSIM-I80 and the NGSIM-US101 dataset during uncongested periods.
The number of training and testing trajectories used for the source target domain are:
Source Domain (NGSIM-I80): 3602 training trajectories, 1796 testing trajectories.
Target Domain (NGSIM-US101): 3897 training trajectories, 3902 testing trajectories.

\subsubsection{Case Study 2 - Cross-period Transfer Learning}
We analyze the effects of latent space domain adaptation on the temporal domain by using trajectory data from the NGSIM-I80 dataset at different time settings, i.e., the uncongested period (4:00pm - 4:15pm) and the congested period (5:00pm - 5:15pm).
The number of training and testing trajectories used for the source target domain are:
Source Domain (4:00pm - 4:15pm): 4104 training trajectories, 2050 testing trajectories.
Target Domain (5:00pm - 5:15pm): 4211 training trajectories, 4216 testing trajectories.

\subsection{Metrics}
In this paper, we report our results in terms of the Root Mean Squared Error (RMSE) of the predicted trajectory in the future (5 second prediction horizon), as follows:
\begin{equation}
\mathrm{RMSE}\,(Y_i, \hat{Y}_i) = \sqrt{\frac{1}{M} \sum_{i=1}^{N} (Y_i - \hat{Y}_i)^2}
\end{equation}
where $M$ is the number of trajectories used for testing.

\definecolor{Gray}{rgb}{0.498,0.498,0.498}
\definecolor{Black}{rgb}{0,0,0}
\begin{table*}[t]
\caption{CROSS-PERIOD STUDY: COMPARISON BETWEEN RMSE OF THE PROPOSED DOMAIN ADAPTED FRAMEWORK AND BENCHMARK MODELS ON SOURCE DOMAIN (4:00PM - 4:15PM). RMSE IS RECORDED IN METERS.}
\label{t3}
\centering
\begin{tblr}{
  cells = {c},
  cell{1}{1} = {c=7}{},
  hline{1} = {-}{},
  hline{2-3,9-10} = {-}{Black},
  vline{7} = {2-9}{},
  vline{6} = {2-9}{}
}
\textbf{Source Domain (4:00PM - 4:15PM)}    &        &                 &          &             &                              &                                 \\
\textbf{Prediction Horizon (s)} & LSTM  & GRIP            & CNN-LSTM & Transformer & {
  GCN-
  \\Transformer
  } & {
  GCN-
  \\Transformer (DA)
  } \\
\textbf{0}                      & 4.9684 & \textbf{2.3192} & 3.4869   & 3.0652      & 2.7127                       & 2.4134                          \\
\textbf{1}                      & 5.6588 & 2.9427          & 3.7983   & 3.8337      & 3.2152                       & \textbf{2.5832}                 \\
\textbf{2}                      & 5.7167 & 3.2330          & 3.5458   & 3.6608      & 3.2735                       & \textbf{2.5559}                 \\
\textbf{3}                      & 6.0208 & 3.8676          & 3.6333   & 4.2602      & 3.4704                       & \textbf{2.6969}                 \\
\textbf{4}                      & 6.9212 & 4.9818          & 4.2042   & 4.9892      & 4.1476                       & \textbf{3.4343}                 \\
\textbf{5}                      & 7.5018 & 5.6497          & 4.7322   & 5.1819      & 4.2431                       & \textbf{3.6003}                 \\
\textbf{Average}                & 6.1313 & 3.8323          & 3.9001   & 4.1658      & 3.5104                       & \textbf{2.8807}                 
\end{tblr}
\end{table*}

\begin{table*}[t]
\caption{CROSS-PERIOD STUDY: COMPARISON BETWEEN RMSE OF THE PROPOSED DOMAIN ADAPTED FRAMEWORK AND BENCHMARK MODELS ON TARGET DOMAIN (5:00PM - 5:15PM). RMSE IS RECORDED IN METERS.}
\label{t4}
\centering
\begin{tblr}{
  cells = {c},
  cell{1}{1} = {c=7}{},
  hline{1} = {-}{},
  hline{2-3,9-10} = {-}{Black},
  vline{6} = {2-9}{},
  vline{7} = {2-9}{}
}
\textbf{Target Domain (5:00PM - 5:15PM)}    &        &        &          &             &                              &                                 \\
\textbf{Prediction Horizon (s)} & LSTM  & GRIP   & CNN-LSTM & Transformer & {
  GCN-
  \\Transformer
  } & {
  GCN-
  \\Transformer (DA)
  } \\
\textbf{0}                      & 6.2607 & 3.1504 & 3.9911   & 3.1710      & 3.1008                       & \textbf{2.7796}                 \\
\textbf{1}                      & 6.7162 & 3.5462 & 4.1113   & 3.5017      & 3.4763                       & \textbf{2.6873}                 \\
\textbf{2}                      & 7.1843 & 4.3535 & 4.2000   & 3.6754      & 3.5605                       & \textbf{2.7037}                 \\
\textbf{3}                      & 7.6275 & 5.1267 & 4.2909   & 4.1328      & 3.8767                       & \textbf{2.9915}                 \\
\textbf{4}                      & 8.1208 & 5.7987 & 4.5580   & 4.6479      & 4.1847                       & \textbf{3.3595}                 \\
\textbf{5}                      & 8.7161 & 6.5414 & 4.8767   & 4.6717      & 4.2231                       & \textbf{3.4969}                 \\
\textbf{Average}                & 7.4376 & 4.7528 & 4.3393   & 3.9667      & 3.7370                       & \textbf{3.0031}                 
\end{tblr}
\end{table*}

\subsection{Benchmark Models}
In this paper, we compare the Domain Adapted Graph Embedded Transformer with the following benchmark models from reviewed literature along with a variant of the proposed Graph Embedded Transformer without domain adaptation:
\begin{enumerate}
    \item \textbf{LSTM}: Vanilla LSTM model \cite{12}.
    \item \textbf{GRIP}: GRIP model \cite{3} integrating GCN and LSTM.
    \item \textbf{CNN-LSTM}: Sequence-to-sequence model \cite{cnnlstm} integrating CNN and LSTM.
    \item \textbf{Transformer}: Vanilla Transformer model \cite{13}.
    \item \textbf{GCN-Transformer}: Baseline of proposed model with GCN and Transformer (no domain adaptation).
\end{enumerate}

\section{Results}
Table \ref{t1} and Table \ref{t2} show the accuracy (in terms of RMSE) of the multi-agent trajectory predictions for the cross-city case study in both the source and target domain, respectively. Table \ref{t3} and Table \ref{t4} show the accuracy of the multi-agent trajectory predictions for the cross-period case study in both source and target domain, respectively.
The results show that integration of the cross-domain adaptation strategy on top of our proposed Graph Embedded Transformer model architecture achieves overall improvements on the prediction accuracy (in RMSEs) over the benchmark models, for both the cross-city and cross-period transfer learning scenarios. In particular, our proposed model with cross-domain adaptation strategy showed increasing improvements in percentages over the benchmarks, as the prediction horizon lengthens. This is due to capability of the self and cross attention mechanism of the Transformer encoder-decoder modules in modelling long dependencies, which play an important role when more input features and temporal patterns are accessible to the decoder module as prediction horizon grows. 

More importantly, the results show that incorporating GCN as the input embedding model of our proposed Graph Embedded Transformer achieved a better prediction accuracy compared to the benchmark models without graph neural network. This is attributed to the competency of the GCN in learning non-Euclidean topological information \cite{3}. The extracted graph feature embeddings thus enhanced the capability of the proposed model in modelling complex spatial interactions between vehicular agents in a traffic scene. Notably, our GCN-enabled model outperforms CNN-LSTM which lacks the ability to capture a global graph-level representation of the system-wide traffic network.

In the cross-city case study, our proposed model outperformed the benchmark models after incorporating the domain adaptation strategy. We attribute this gain in performance to the domain adaptation strategy in learning attention feature representation that is adaptable to traffic domains with statistical discrepancy, especially in the cross-city scenario where the I80 (East-West Freeway) and US101 (North-South Freeway) exhibit both distinct spatial and temporal characteristics. With the the integration of domain adversarial training, our proposed model and domain adaptation strategy showed a RMSE improvement of 20.81\% and 21.58\% over the Graph Embedded Transformer baseline on the source and target domains, respectively. The results thus highlights the necessity of the proposed domain adversarial training in adapting the learning of the attention feature representation on complex multimodal traffic system.

In the cross-period case study, however, our proposed model outperforms the benchmark model even without domain adaptation. These initial RMSE improvements in both the time domains are due to the competency of the Transformer model in modelling complex temporal patterns on unimodal traffic system with uniform spatial characteristic, considering that the models are evaluated on a single location, I80 (East-West Freeway).
A further integration of domain adversarial training on the baseline allows the attention features to generalize better across time periods, which we observes a further RMSE improvement of 17.94\% and 19.64\% over the Graph Embedded Transformer baseline on the source and target domains, respectively.


\section{CONCLUSION}
In this study, we introduce a graph-based Transformer network known as the Graph Embedded Transformer with a domain adaptation framework for predicting multi-agent vehicle trajectories of highway traffic scene in encompassing adaptation in different locations and time periods. Agents in the traffic network encompasses vehicles in an urban traffic. Our domain adaptation approach capitalizes on the latent features learning from graph encoding of the GCNs along with self and cross attention features of the Transformer's encoder and decoder modules. 
Our experimental findings, conducted on the NGSIM-I80 and NGSIM-US101 datasets, reveal that the proposed domain adaptation strategy using attention features of the Graph Embedded Transformer significantly improves cross-domain trajectory prediction. Incorporating domain adaptation 
to our proposed model observes improvement in performances on both source and target domains, over the benchmarks without domain adaptation. 

\bibliographystyle{IEEEtran}
\bibliography{1,2,3,4,5,6,7,8,9,10,11,12,13,14,15,16,17,18,19,20,21,22,23,24,gcn1,gcn2,gcn3,gcn4,tf1,cnnlstm,lstmori,transori,Chai,Ganin} 

\end{document}